\documentclass[11pt]{article}

\usepackage[final]{acl}

\usepackage{times}
\usepackage{latexsym}

\usepackage[T1]{fontenc}

\usepackage[utf8]{inputenc}

\usepackage{microtype}

\usepackage{inconsolata}

\usepackage{graphicx}

\title{The Environmental Impacts of Language Model Training Keep Rising\\ 
Now is the Time to Catch Impacts on the Rebound}

\author{Cl\'ement Morand \and Aur\'elie N\'ev\'eol \\
  Université Paris-Saclay, CNRS, LISN \\
  \texttt{name.surname@lisn.fr} \\\And
  Anne-Laure Ligozat \\
  ensIIE \\
  Université Paris-Saclay, CNRS, LISN\\
  \texttt{name.surname@lisn.fr} \\}

\usepackage{subcaption}
\usepackage{multirow}
\usepackage{makecell}
\usepackage{booktabs}
\usepackage{xurl}

\usepackage[acronyms,toc]{glossaries}
\makenoidxglossaries
\newacronym{nlp}{NLP}{Natural Language Processing}
\newacronym{llm}{LLMs}{Large Language Models}
\newacronym{ml}{ML}{Machine Learning}
\newacronym{DL}{DL}{Deep Learning}
\newacronym{ICT}{ICT}{Information and Communication Technology}
\newacronym{PUE}{PUE}{Power Usage Effectiveness}
\newacronym{TDP}{TDP}{Thermal Design Power}
\newacronym{AI}{AI}{Artificial Intelligence}
\newacronym{LCA}{LCA}{Life Cycle Assessment}
\newacronym{CI}{CI}{Carbon Intensity}
\newacronym{ADPe}{ADPe}{Abiotic Depletion Potential for elements}
\newacronym{GWP}{GWP}{Global Warming Potential}

\newcommand{\COtwo}{CO\textsubscript{2}~eq}
\newcommand{\Sbe}{Sb~eq}

\begin{document}

\maketitle

\begin{abstract}
Recent Machine Learning (ML) approaches have shown increased performance on benchmarks at the cost of escalating compute demands.
Hardware, algorithmic and carbon optimizations have been proposed to curb energy use and environmental impacts.
We estimate the environmental impacts associated with training 
models documented in the Epoch AI database
over the last decade, with a particular focus on 
impacts associated with Large Language Models and the hardware used to train them. 
We find that 
energy use and environmental impacts associated with training ML models have increased exponentially, even when considering impact reduction strategies such as using less carbon intensive electricity mixes or more efficient hardware.
Optimization strategies do not mitigate the impacts induced by model training, suggesting rebound effect.
We show that the impacts of hardware must be considered over the entire life cycle rather than the sole use phase in order to avoid impact shifting.
Our study demonstrates that increasing efficiency alone does not ensure sustainability. 
There is an urgent need to systematically integrate environmental impacts in NLP evaluation practices to better inform the community and support the use of impact as a feature in research planning and decision making.
\end{abstract}

\section{Introduction}

The increasing ubiquity of \gls{llm} is fueling debates in the \gls{nlp} community who is questioning the risks and potential associated with global LLM use~\cite{Bender2021parrots,scialom-etal-2022-fine,groeneveld-etal-2024-olmo,Varoquaux2025bigger}. 

NLP evaluation uses community-developed tools including benchmarks and standardized performance metrics that aim for easy and reproducible comparison between models - as shown in leaderboards. 
Standard \gls{nlp} evaluation methods over-emphasize task related performance at the expense of user oriented metrics~\citep{Ethayarajh2020utility} and real world impact evaluation of \gls{nlp} applications~\citep{Reiter2025evaluate}. 
The community has identified a need to rethink progress and evaluation practices. We believe that the systematic assessment of compute requirements and associated environmental impacts is a fundamental way forward. 
The impacts of \gls{ml}, including \gls{llm}, come from many sources including model development, training and use~\citep{Ligozat2021unraveling,Wu2022sustainable}.
Although model inference has received much attention with the wide use of \gls{llm}, model training impacts are substantial compared to inferences~\citep{Wu2022sustainable,berthelot2024generative} and need careful assessment.  

Research is underway to understand and develop assessment methods for the impacts of carbon emissions~\citep{Strubell2019energy}, water usage~\citep{li2025makingaithirstyuncovering} and metallic resource depletion~\citep{Morand2024MLCA,berthelot2024generative}.
However these studies are limited to individual models cases and, to our knowledge, no comprehensive analysis of LLM training impact trends is available.

Strategies have been developed to optimize the energy use of models training. \textit{Hardware optimizations} leverage more recent energy-efficient hardware. \textit{Algorithmic optimizations} adjust software to offer the same task performance using less compute. \textit{Carbon optimizations} displace computation towards less carbon-intensive electricity (e.g., renewable energy). Major \gls{AI} companies have claimed these strategies would mitigate \citep{wu2024efficiency} or reduce the carbon footprint of \gls{ml} \citep{Patterson2022plateau}.

These claims overlook two important points. First, frequent hardware upgrades reduce the carbon footprint from the use phase at the expense of the production and end-of-life phases. They also increase metallic resource depletion, limiting the overall benefit. This phenomenon is called \emph{impact shifting}. Second, optimization may result in smaller-than-expected reductions, or even increase the overall environmental impacts because of \emph{rebound effect}, a prevalent phenomenon in \gls{ICT}~\citep{Gossart2015rebound,Bol2021Moore}.

In this study, we address the following research question: "Do optimization strategies lead to lower impact for \gls{llm} model training over time?". 

In section 2, we review related work. In section 3 we detail the impact assessment method we combine with data from GPU manufacturers and the Epoch AI database to  examine (in section 4) the environmental trajectory of ML model training over time. Finally, in section 5, we investigate the efficacy of carbon, hardware, and software optimizations in mitigating environmental impacts. Our analysis reveals that the increase in compute demand has outweighed optimization gains. Overall, optimizations are correlated with larger models, consistent with rebound effect. As carbon optimization strategies are insufficient to curb the growth in the environmental impacts of model training, we advocate integrating environmental impact into NLP evaluations to provide actionable data to guide community decision making.

\section{Related Work: Measuring the Environmental Impact of AI training}
\label{sec:sota}

After the high level of carbon emissions associated with training \gls{nlp} models was reported~\citep{Strubell2019energy}, researchers stated the need for "Green AI"~\citep{Schwartz2020green}, soon structured as an entire research field focused on lowering impacts~\citep{treviso-etal-2023-efficient, Verdecchia2023systematic}. 

\paragraph{Tools to Measure the Impact of an Algorithm} Methods and tools to assess the energy use of \gls{AI} activities have been developed, 
based on energy use measures (e.g., {\em CarbonTracker} \citep{anthony2020carbontracker}) or estimations (e.g., {\em Green Algorithms} \citep{Lannelongue2021green}). 
Energy use assessments are complemented with data center energy efficiency indicators 
such as \gls{PUE}~\citep{Avelar2012pue}, and information on the carbon intensity of local electricity mixes to convert energy use into carbon footprint assessments. 
Reviews have compared tools to assess the energy use and associated carbon footprint of training language models and other \gls{AI} activities~\citep{Bouza2023estimate,Jay2023experimental,Bannour2021evaluating}. 
Additional efforts address the water usage associated with hardware energy use by using the \textit{Water Usage Efficiency} (WUE) of data centers~\citep{li2025makingaithirstyuncovering}.

\gls{ml} related impacts are also incurred by model development (including dataset curation or architecture search) and use, with  environmental impacts coming from each life cycle phase of hardware used~\citep{Ligozat2021unraveling}.
MLCA~\citep{Morand2024MLCA} leveraged an existing method~\cite{Luccioni2022estimating} to assess environmental impacts associated with ML training or inferences, including carbon footprint, energy demand and metallic resource depletion over the production and use phase of the hardware.
\citet{morrison2025holistically} build on \citet{Luccioni2022estimating} and \citet{li2025makingaithirstyuncovering} to assess carbon footprint and water use of model training, including development and simulated inference. However, environmental assessments that scale beyond individual models are still needed.

\paragraph{Drivers of AI Impact Growth Remain Unclear}

Technical reports from industry and academia suggest that the environmental impacts associated with AI have been rising at the global level~\cite{Google2024environmentalreport,Meta2024environmentalreport,Shehabi2024USdatacenterreport}. 
\citet{devries2023growing} discussed the potential growth of \gls{ml} with the recent surge in demand for freely accessible \gls{llm}. The \citet{IEA2024electricity} stated that 
energy use by AI and cryptocurrencies could double by 2026.
\citet{Masanet2020recalibrating} have shown that despite an exponential growth in computation demand between 2012 and 2018, the total energy use of data centers had only increased by around 5\% thanks to a shift toward more energy efficient large scale data centers. While this relative decoupling is no longer valid \citep{IEA2025EnergyandAI}, we lack consolidated information on the global trend for environmental impacts of \gls{AI}. Information is needed on the impact trend of each step, including training. 

\paragraph{Diverging Trends Need to be Analyzed for LLM Training} 

Optimization techniques have been used to produce models reaching a given level of performance at a reduced training carbon footprint \citep{Wu2022sustainable, Patterson2022plateau}.
The energy efficiency of  graphics cards has increased exponentially, leading to the claim that frequent hardware updates will shrink the carbon footprint of training models \citep{Patterson2022plateau}.
In spite of algorithmic optimization, \citet{sevilla2022compute} and \citet{Thompson2023Computational} report an exponential increase in compute requirements to train models.
Compute requirements are increasing at the same time as energy efficiency, leading to unclear trends on model training impacts. 
Optimizations are often absorbed by the growth of the \gls{ICT} sector~\cite{Bol2021Moore,Gossart2015rebound}.
Is this also true for \gls{llm} training?
\citet{Luccioni2025rebound} warn against risks of rebound effect in AI. 
A \emph{producer rebound} effect \citep{Coroama2019rebound} could be occurring, where efficiency increases fuel the creation of even larger models thereby lowering or even canceling the potential impact reduction.

Overall, previous work has led to the development of methods and tools for assessing the impact of specific \gls{ml} algorithms. 
Reports suggests that the overall environmental impact of \gls{AI}
is rising. However, mechanisms behind such an increase remain undocumented.   
Deployment of optimizations concurring with growing compute demand could correspond with rebound effect in the case of model training. 
Herein, we present a comprehensive study of the impact of \gls{ml} hardware and \gls{ml} model training over a decade, including hardware production and investigating rebound effect.

\section{Method: Environmental Impact Assessment}
\label{sec:methods}

In this article, we question whether optimization strategies lead to lower training impact.
This Section presents the environmental assessment method used in Section~\ref{sec:trends} to plot impact trends over time. 
The code and information used in the experiments are available on zenodo~: \url{https://doi.org/10.5281/zenodo.15310868}.

\subsection{Data sources}
\label{subsec:Machine-learning}

Studies on \gls{ml} models training are conducted using the Epoch AI Notable systems database \citep{EpochNotableModels2024}. Epoch AI is the most comprehensive database on \gls{ml} systems to our knowledge. It gathers extensive information on a large variety of \emph{notable}\footnote{Notable systems are ``models that have advanced the state of the art, had a large influence in the field’s history, or had a large impact within the world'', per \url{https://epochai.org/data/notable-ai-models-documentation}.} \gls{ml} systems. 
The database mostly comprises data gathered from papers/technical reports presenting the models and used to yield estimates with varying levels of confidence, but some parameters cannot be reliably estimated. 
Information on graphics cards characteristics is based on a dataset
gathering manufacturer information~\citep{Morand2025dataset}. Both
data sources are available under Creative Commons Attribution
license. Additional details to support replication are available in Appendix~\ref{app:replication}.

\subsection{General Methodology for Environmental Impact Assessment}
\label{subsec:metrics}

To address our research question, we used a methodology to assess the impacts of \gls{ml} system training including hardware life cycle. This process requires detailed information on hardware characteristics and specific training parameters, such as training duration.

\paragraph{Assessing Multiple Impacts Over Hardware Life-Cycle}

The MLCA tool~\citep{Morand2024MLCA} is chosen to perform environmental assessments. It uses attributional LCA, a methodology allowing to account for multiple impacts over the life cycle of the used hardware. It provides carbon footprint and metallic resource depletion assessments, based on open-source \gls{LCA} information from~\citet{Simon2024boaviztapi}\footnote{
Water use and toxicity to human and non-human life are also important sources of impact for \gls{ICT} hardware and services. Recent open quality information on toxicity and water use of ICT equipment and data center facilities is missing~\citep{li2025makingaithirstyuncovering,Lei2022wue}. End-of-life information is also lacking~\citep{Ficher2025eol}.}.  

\begin{figure}[ht]
    \centering
    \includegraphics[width=\linewidth]{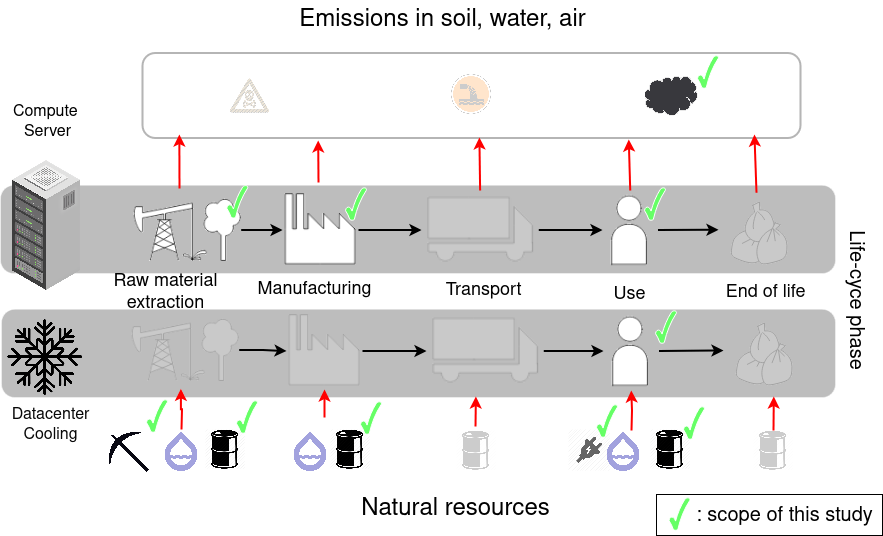}
    \caption{Scope of environmental assessments of \gls{ml} training in this study. Pictograms represent fossil fuels (barrel), water (droplet),  metals (pickaxe) and electricity (plug). Top row icons represent, from left to right, emissions in soil, water and air.}
    \label{fig:shema-scope}
\end{figure}

Figure~\ref{fig:shema-scope} presents the scope of our study within the life cycle of hardware used for training ML models. We consider energy use, carbon footprint and metallic resource depletion over server production and usage, and data center cooling use.

Carbon footprint is assessed according to \gls{GWP},\footnote{We use the IPCC standard \gls{GWP}100, corresponding to impact at a time horizon of 100 years.} measured in kg\COtwo{}, for the emissions of greenhouse gases such as carbon dioxide, methane and nitrous oxide~\citep{IPCC2023earth}. 
Metallic resource depletion is assessed through \gls{ADPe}, expressed in \emph{kilograms antimony equivalent} (kg\Sbe{}) \citep{van2020abiotic}, which is based on the quantity of metal (e.g., copper, gold and rare earths) used to produce hardware. The amount of each metal and metal rarity are compared to Antimony, chosen as standard reference, to create and aggregated indicator.

\begin{figure}[ht]
    \centering
    \includegraphics[width=\linewidth]{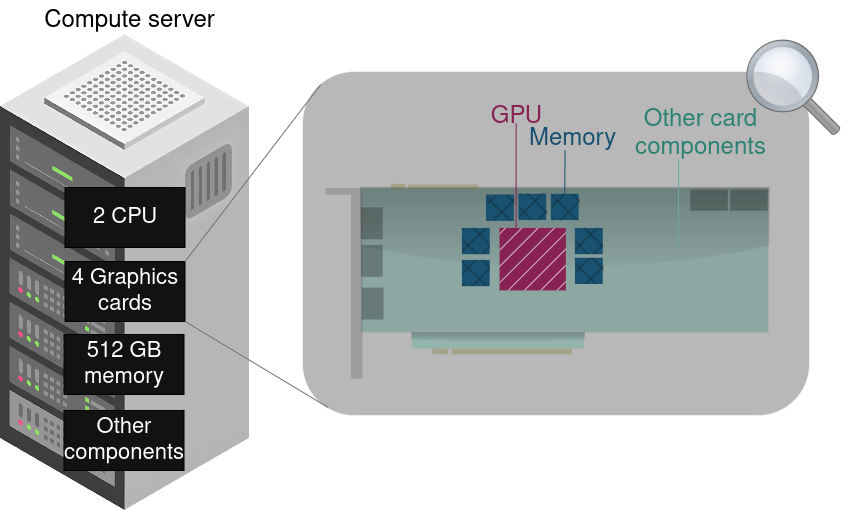}
    \caption{Hardware modeling in MLCA.}
    \label{fig:schema_modelisation_MLCA}
\end{figure}

MLCA is based on a bottom-up modeling of hardware, meaning that hardware is modeled as the sum of its parts as represented in Figure~\ref{fig:schema_modelisation_MLCA}. 
Graphics cards are modeled based on GPU size (in hatched purple in the Figure), on-board memory size (in cross-hatched blue) and constant impact for other card components (in green). 
Their energy use during \gls{ml} training is estimated using their \gls{TDP}\footnote{The \gls{TDP} corresponds to the heat the chip should be able to dissipate to function at maximum load; it is used to approximate the maximum power use of the chip.} and training duration.

\paragraph{Training Duration}

\newcommand{\firstestimate}[0]{\texttt{GPU-h\textsubscript{1}}}
\newcommand{\secondestimate}[0]{\texttt{GPU-h\textsubscript{2} base}}
\newcommand{\secondestimateclean}[0]{\texttt{GPU-h\textsubscript{2}}}

We estimate GPU-hours required for training models using two methods, based on information from the Epoch AI database presented in Table~\ref{tab:data set_specs}. 
The estimates are performed on model entries majorly assessed as reporting high-quality information ($74\%$ of models).

\begin{table*}[ht]
    \centering
    \small
    \begin{tabular}{cc c ccc cccc}\toprule
                             & & & \multicolumn{4}{c}{\firstestimate{}} & \multicolumn{3}{c}{\secondestimateclean{}}   \\\cmidrule(lr){4-7}\cmidrule(lr){8-10}
        \multicolumn{2}{c}{} & models & \thead{training\\duration} & \thead{card\\quantity} & both & \thead{+ card\\model} & \thead{training\\FLOP} & \thead{card\\model} & both  \\
        \midrule
        \multicolumn{2}{c}{Number}               & 897 & 199  & 168  & 131  & 122 & 451  & 266  & 230 \\
        \multicolumn{2}{c}{Coverage (\%)}        & 100 & 22 & 19 & 15 & 14 & 50 & 30 & 26 \\
        \addlinespace
        \multirow{4}{*}{\rotatebox[origin=c]{90}{Confidence}} & Confident   & 234 & 98   & 96   & 78   & 75 & 166  & 138  & 122   \\
                                    & Likely      & 103 & 41   & 34   & 21   & 17 & 77   & 51   & 47  \\
                                    & Speculative & 69  &  21   & 11   & 9    & 7 & 51   & 20   & 16 \\
                                    & Unknown     & 491 & 39   & 27   & 23   & 23 & 157  & 57   & 45 \\\bottomrule
    \end{tabular}
        \caption{Number of entries for each information type in the Epoch AI database. Confidence scores are reported by database authors based on the quality of the Training compute, Parameters, and Training data set size fields.}
    \label{tab:data set_specs}
\end{table*}
For models specifying duration and graphics cards quantity, they can be multiplied to obtain an estimate referred to as  \emph{\firstestimate{}}: 
$$\firstestimate{} = \texttt{training duration} \times \#\texttt{cards}$$
This estimation is the most reliable as it uses information retrieved 
from the papers presenting the models.
However, it only covers $15\%$ of Epoch AI models (see Table~\ref{tab:data set_specs}). 

Another method to estimate the training duration amounts to dividing the training compute requirements (number of FLOP) necessary for model training by the peak compute power of the card used
(in FLOPs, i.e., the number of FLOP the card can perform per second). This value is referred to as \emph{\secondestimate{}}: 
$$\secondestimate{} = {\small\frac{\texttt{training FLOP}}{\texttt{peak card compute power}}}$$
\secondestimate{} should lead to an under-estimation of the number of GPU-hours compared to \firstestimate{}, as the hardware does not always operate at peak performance, for instance due to synchronisation phases between the different nodes when compute cannot be realised.

After checking for consistency between the two estimation methods and excluding benign anomalies, we develop a linear model to correct the underestimation from \secondestimate{} compared to \firstestimate{}. Appendix~\ref{sub-app:duration} details procedures for ensuring consistency and details the linear model. Appendix~\ref{app:sensitivity} explores the sensitivity of the estimated training duration to variations in the used performance ratio, and the influence these variations have on the assessed training carbon footprint and metallic resource depletion.
The obtained model corresponds to using a quasi-constant performance ratio of $\simeq 27\%$.
This leads to our second estimation method referred to as \emph{\secondestimateclean{}}: 
$${\small\secondestimateclean{} = \frac{\texttt{training FLOP}}{\texttt{peak card compute power} \times 0.27}}$$

The final estimation for GPU hours is as follows: we use \firstestimate{} for the 131 models (15\%) where it is available, and use \secondestimateclean{} for another 103 models (11\%), covering $26\%$ of Epoch AI models.
Among these 234 models, 132 (56.4\%) are \gls{nlp} models (indicating at least language or speech modality). 123 (95.3\%) of these models are language models, and the 6 other (4.7\%) are speech models. 26 (20.1\%) \gls{nlp} models are multimodal. We include models in domains other than \gls{nlp} to refine potential trends even if our primary focus is language models.
In total, 30 (12.8\%) of the 234 models indicate a base model, and are thus considered to be fine-tuned or distilled. Training duration is estimated using \firstestimate{} for these models and only accounts for the fine-tuning phase and not the base model training.

\paragraph{Illustration on GPT-4}

We illustrate our methodology for the GPT-4 model, released in March 2023.
Training duration is 
estimated at $57,000,000$ GPU-hours on \textit{NVIDIA A100 SXM4 40 GB} cards based on information in the EpochAI dataset and using \firstestimate{}. 
Training servers are modeled as containing 4 cards, 2 CPUs and 512GB memory (consistently with NVIDIA workstation cards), and facility PUE is estimated at 1.2 (because the model was trained after 2018). 
Energy use impact is computed with average USA energy mix (GPT4 authors are affiliated in the US). Table~\ref{tab:GPT4} presents impacts associated with GPT-4 training estimated using MLCA. For this model, embodied impacts represent 20\% of the total carbon footprint and close to 100\% of metallic resource depletion. Carbon footprint estimation is consistent with previous assessments\footnote{\url{https://medium.com/data-science/the-carbon-footprint-of-gpt-4-d6c676eb21ae} [Last Accessed 16/04/2026]}.

\begin{table}[htbp]
    \centering
    \small
    \begin{tabular}{cccc}\toprule
                 & Energy & \gls{GWP}         & \gls{ADPe} \\
                 & (GWh) & (kt\COtwo{}) & (kg\Sbe{})\\\midrule
        Embodied & - & 3.3  & 300 \\
        Usage    & 27.4 & 10.2 & 2.7 \\
        Infra    & 5.4 &  2.0 & 0.5 \\\addlinespace
        Total    & 32.8  & 15 & 300 \\\bottomrule
    \end{tabular}
    \caption{Estimated production impacts for training GPT-4: impacts related to the energy use of servers (row \textit{Usage}),  infrastructure (row \textit{Infra}) and embodied impacts (row \textit{Embodied}) are presented.
    }
    \label{tab:GPT4}
\end{table}

\paragraph{Simulating Carbon Optimization Strategies}
\label{subsec:carbon_reduction}
The study considers different scenarios for reducing \gls{CI} in energy mixes over time, including shifting compute locations and decarbonizing electricity for data centers. Each scenario involves a continuous reduction in  \gls{CI}, starting from 2019, with reductions of up to 25\% per year onward. The \gls{CI} of the mix used for training a model is thus multiplied by $(1-\texttt{ratio})^{\texttt{n}}$ where n stand for the number of years since 2019 at release date of the system.

\section{Results: Trends in \gls{ml} Model Training}
\label{sec:trends}

Now that we have data and a method for environmental assessments, we
aim at answering the following questions: How have graphics cards
production impacts evolved? How have model training impacts evolved?

\paragraph{Production impacts of the used cards}

\begin{figure*}
\begin{subfigure}{.45\linewidth}
        \centering
    \includegraphics[width=0.95\columnwidth]{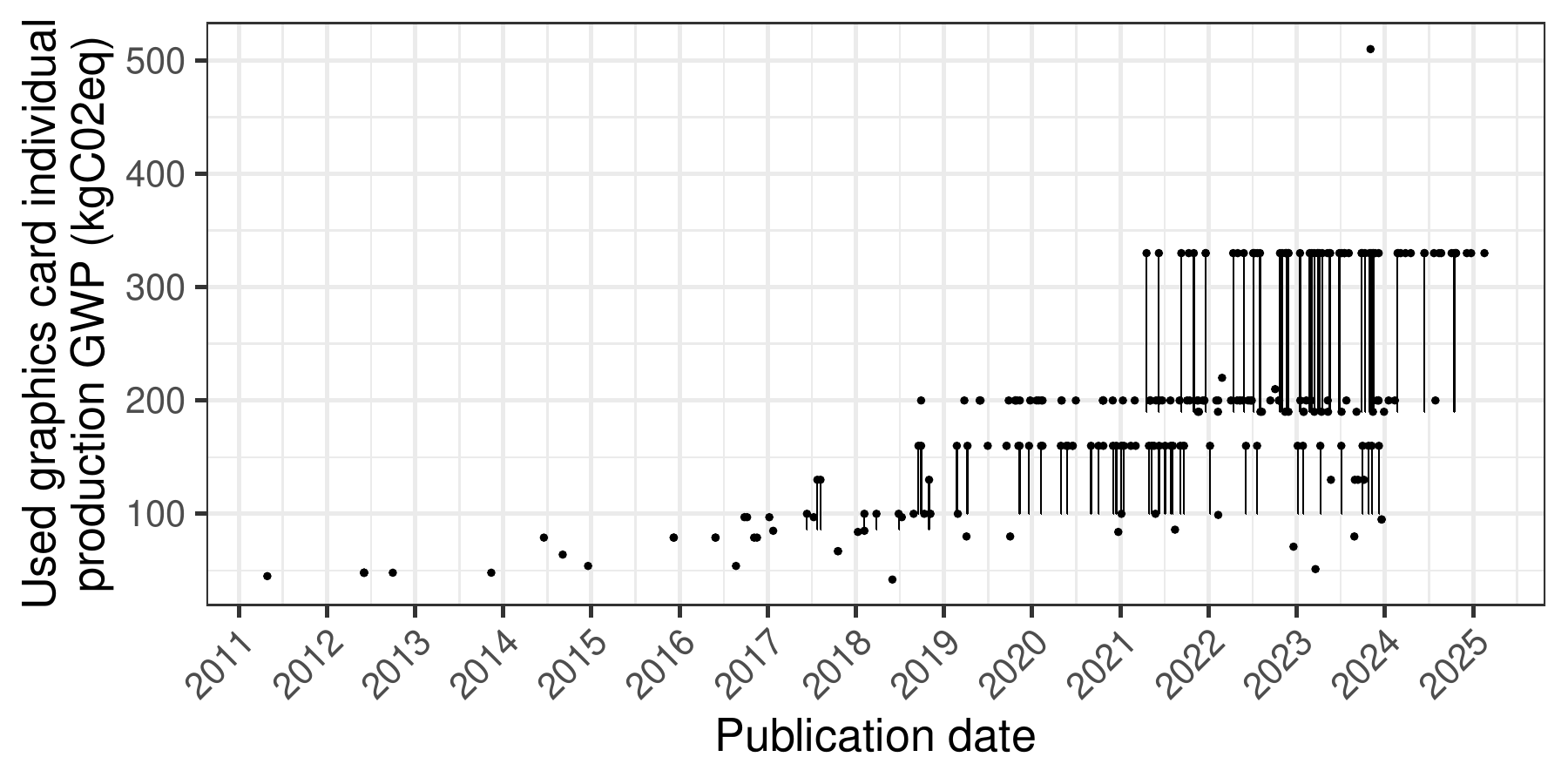}
    \subcaption{\gls{GWP}}
    \label{fig:production_GWP_used_GPU}
\end{subfigure}
\hfill
\begin{subfigure}{.45\linewidth}
    \centering
    \includegraphics[width=0.95\columnwidth]{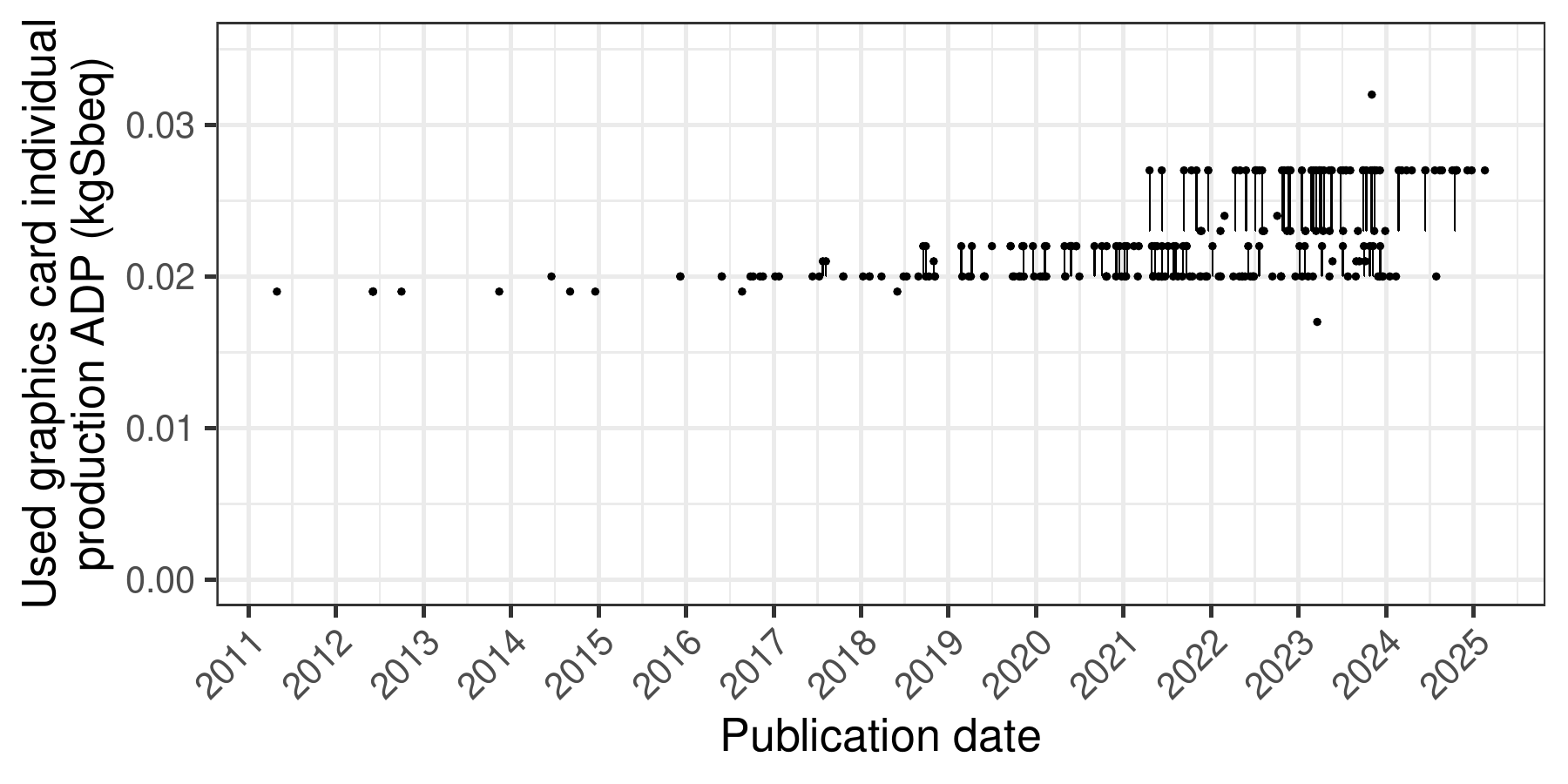}
    \caption{\gls{ADPe}}
    \label{fig:production_ADP_used_gpu}
\end{subfigure}
    \caption{Evolution of the production impacts of the graphics cards (assessed using the MLCA tool) used for training \gls{ml} systems in the Epoch AI database. Value intervals correspond to cases of ambiguous card names.
    }
    \label{fig:production_used_gpu}
\end{figure*}

More recent models are trained on newer cards. The new cards are more technologically advanced, allowing for increased compute capacities and energy efficiency. As shown in Figure~\ref{fig:production_used_gpu}, this however leads to increases in the production impacts of graphics cards used to train \gls{ml} models. 

\begin{figure}[ht]
    \centering
    \includegraphics[width=0.95\columnwidth]{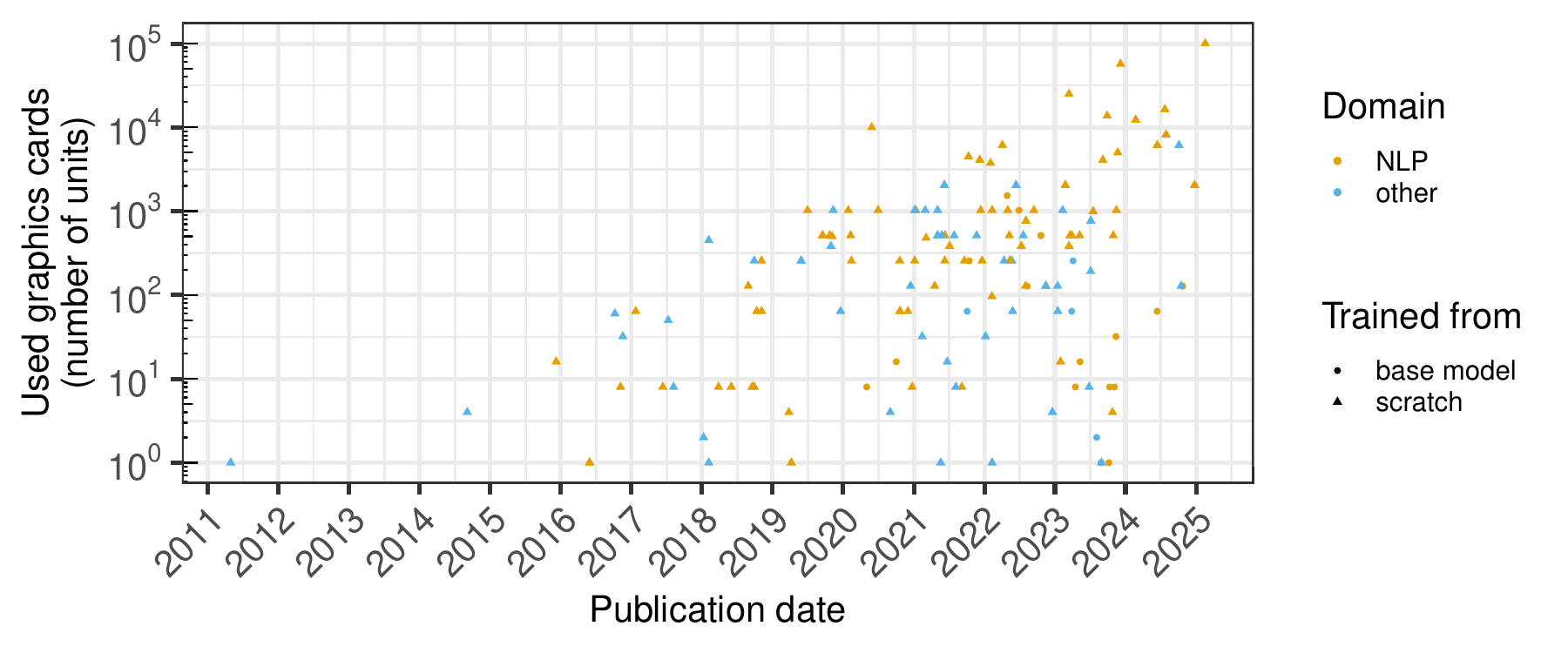}
    \caption{Number of graphics cards used for training models in Epoch AI over time. Y scale is logarithmic so a linear trend corresponds to an exponential growth.}
    \label{fig:hardware_quantity}
\end{figure}

Figure~\ref{fig:hardware_quantity} shows that hardware quantity has furthermore increased exponentially, even if some models are still trained using only a few cards. More graphic cards with higher production impacts are used, suggesting growing environmental impacts. Both production impact and energy use during usage need to be addressed.

\paragraph{Energy use of Model Training}

\begin{figure}[ht]
    \centering
    \includegraphics[width=.95\columnwidth]{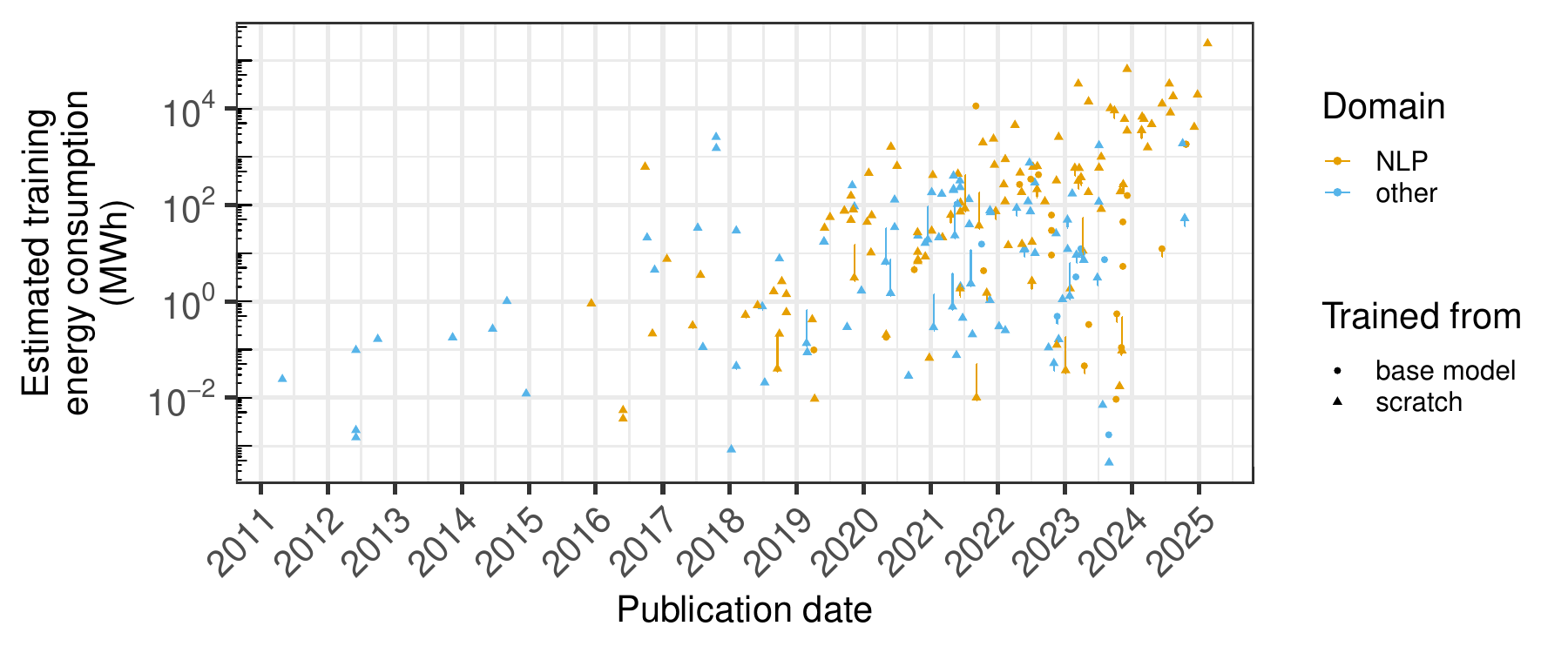}
    \caption{Energy use of training over time. Value intervals account for ambiguous card names. Y scale is logarithmic so a linear trend means exponential growth.
    }
    \label{fig:dynamic_energy_consumption}
\end{figure}

Figure~\ref{fig:dynamic_energy_consumption} presents the energy use
associated with model training estimated using MLCA (see
Section~\ref{subsec:metrics}). Energy use has increased exponentially
over time. Appendix~\ref{app:regression} details regression analysis
for the assessed trends in energy use and environmental impacts. While Low energy use models are still produced in recent years, the number of models released over time has increased overall.

\paragraph{Environmental Impacts of Model Training.}

\begin{figure*}[ht]
\begin{subfigure}{.45\linewidth}
    \centering
    \includegraphics[width=.95\columnwidth]{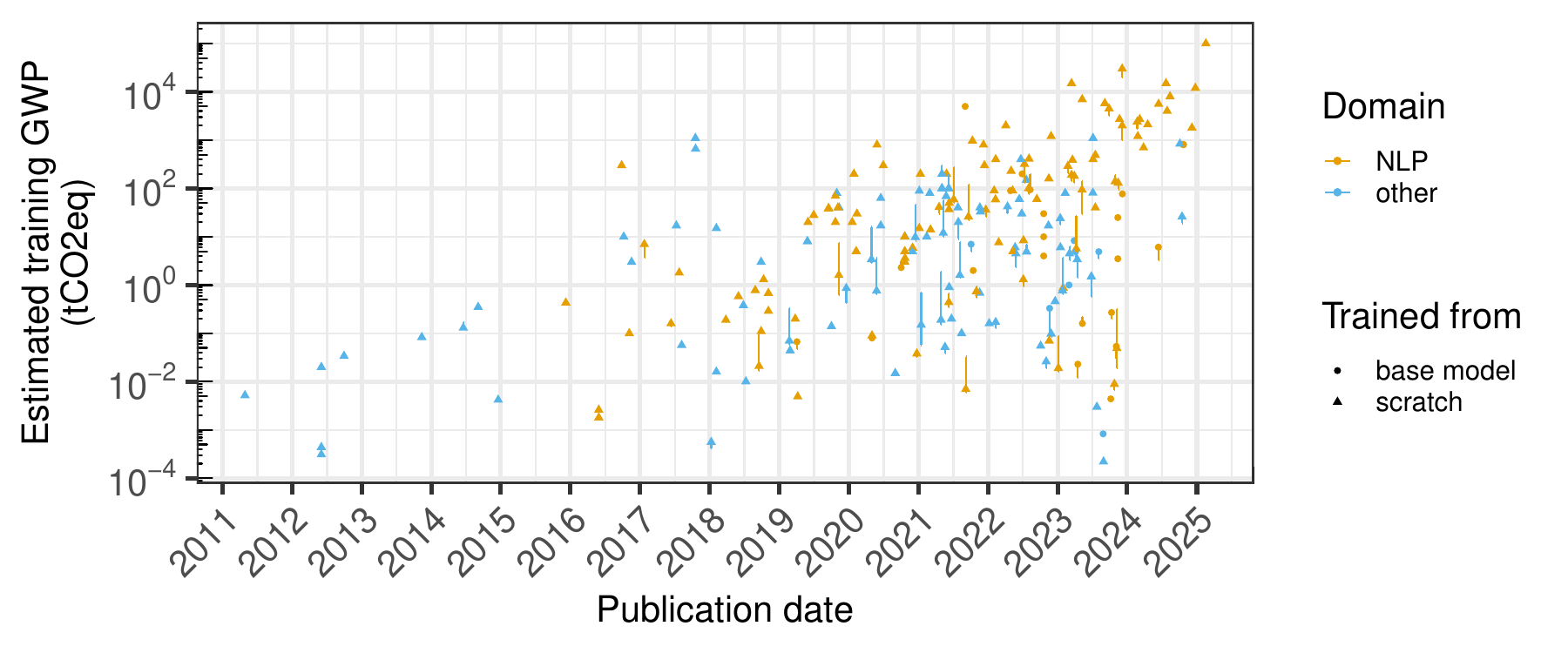}
    \subcaption{\gls{GWP}}
    \label{fig:GWP_training}
\end{subfigure}
\hfill
\begin{subfigure}{.45\linewidth}
    \centering
    \includegraphics[width=.95\columnwidth]{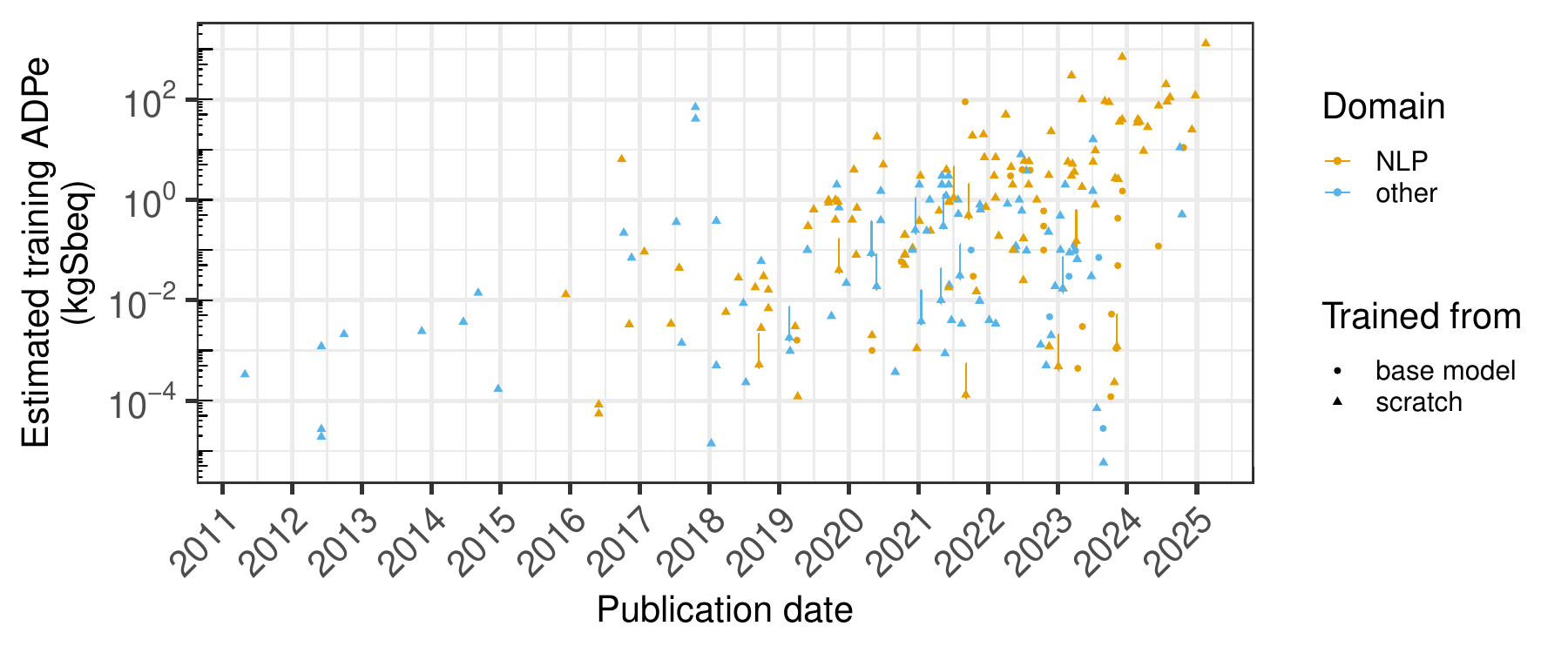}
    \subcaption{\gls{ADPe}}
    \label{fig:ADP_training}
\end{subfigure}
 \caption{Environmental impacts in \gls{GWP} (a) and \gls{ADPe} (b) of training \gls{ml} models over time. Value intervals account for ambiguous card names and model producers from multiple countries. Y scales are logarithmic so a linear trend corresponds to an exponential growth.}
 \label{fig:footprint_training}
\end{figure*}

Figure~\ref{fig:footprint_training} presents the estimated training impacts in terms of \gls{GWP} (Figure~\ref{fig:GWP_training}) and \gls{ADPe} (Figure~\ref{fig:ADP_training}). Both environmental indicators have increased exponentially between 2012 and 2025. \gls{nlp} appear representative of other domains. Additional disaggregated experiments indicate that language, vision, and multimodal models (to a lesser extent) all exhibit an exponential growth trend.

Table~\ref{tab:share_adp} summarizes the distribution of the shares of embodied impacts.
\gls{ADPe} comes close to exclusively from hardware production. 
The share of embodied \gls{GWP} are mostly estimated around a third the total, which is consistent with previous assessments in the literature~\citep{Gupta2020chasing,wu2024efficiency,Luccioni2022estimating}. Variation mostly comes from differences in the \gls{CI} of the electricity mixes of countries in which models are trained. Models trained in countries with a low \gls{CI} for their electricity use such as France will present larger shares of embodied impacts than if trained in another country such as the USA.
Embodied impacts represent a significant share of total impacts for both indicators. Thus, solely reducing the energy impacts will not be sufficient to solve the environmental impacts of \gls{ml} models training.

\begin{table}[ht]
    \centering
    \small
    \begin{tabular}{@{}ccccccc@{}}\toprule
               & Min   &   Q1 &     Q2   &     Mean  &    Q3   &     Max \\
                \midrule
            \gls{ADPe} & 89 & 100 & 100 & 99 & 100 & 100 \\
            \gls{GWP} & 12 & 19  & 23  & 23 & 25  & 58 \\\bottomrule
    \end{tabular}
    \caption{Share (in percent) of embodied impacts on the total impacts associated with training models.}
    \label{tab:share_adp}
\end{table}

\section{Result Analysis: Why do Training Impacts Rise in spite of Optimization?}\label{sec:optimization}
Rising impact trends may appear surprising as optimisation strategies are being deployed. In this section, we investigate what role hardware and software optimization played in these trends and whether carbon optimization can mitigate impact. 

\subsection{Hardware optimization}

\textit{Compute efficiency} is computed by dividing the peak performance of cards (maximum compute power in Single, Double, Half and Tensor precision) by 
\gls{TDP}. It corresponds to the number of operations cards can perform per second per Watt. It is expressed in FLOPs/W, and larger values are better.

\begin{figure}[ht]
    \centering
    \includegraphics[width=0.9\columnwidth]{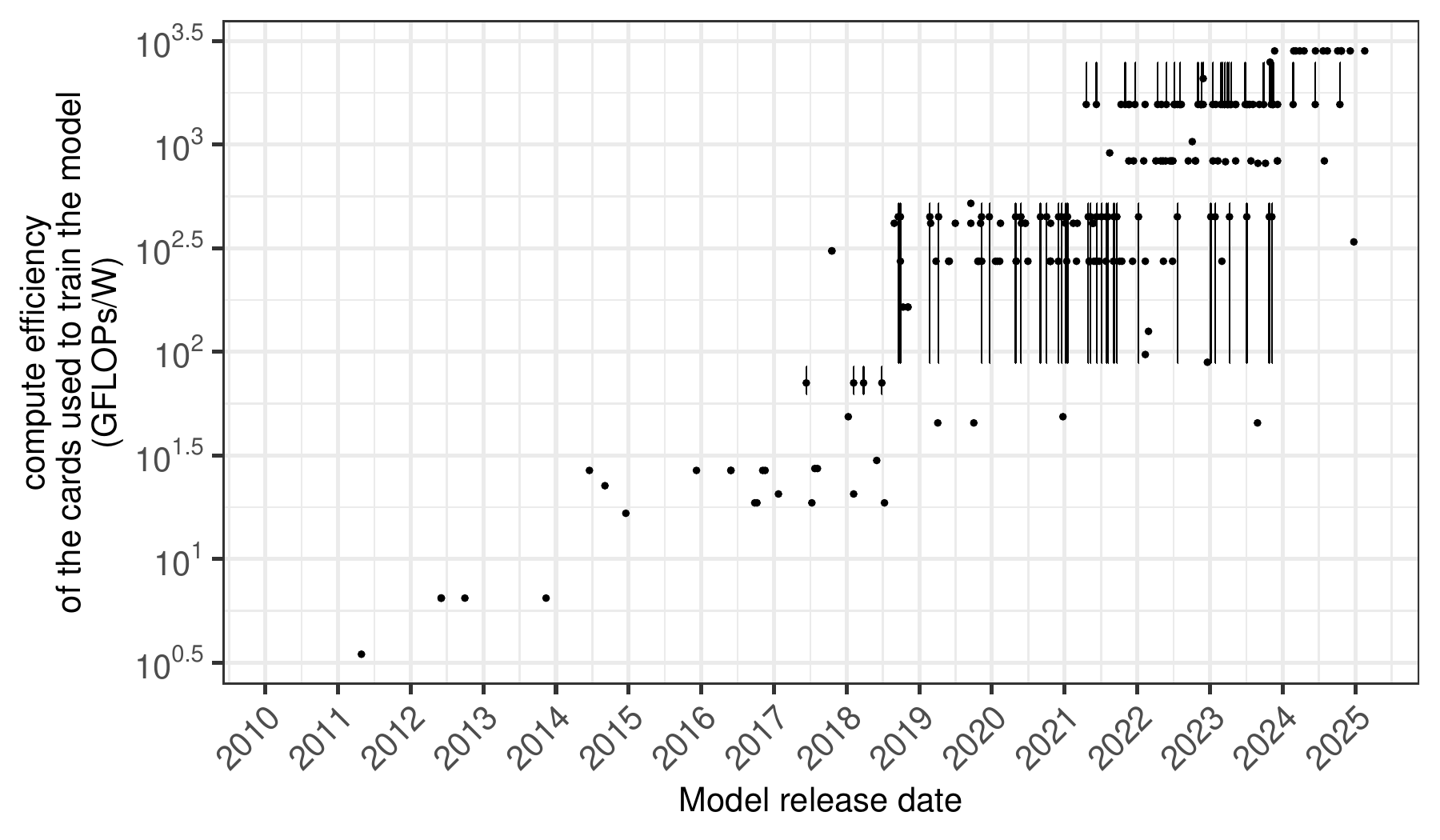}
    \caption{Evolution of the compute efficiency of the graphics cards used to train models in the Epoch AI database (2013-2025). Higher is better. Value intervals account for ambiguous card name}
    \label{fig:evol_compute_efficiency}
\end{figure}

\begin{figure}[ht]
    \centering
    \includegraphics[width=0.9\columnwidth]{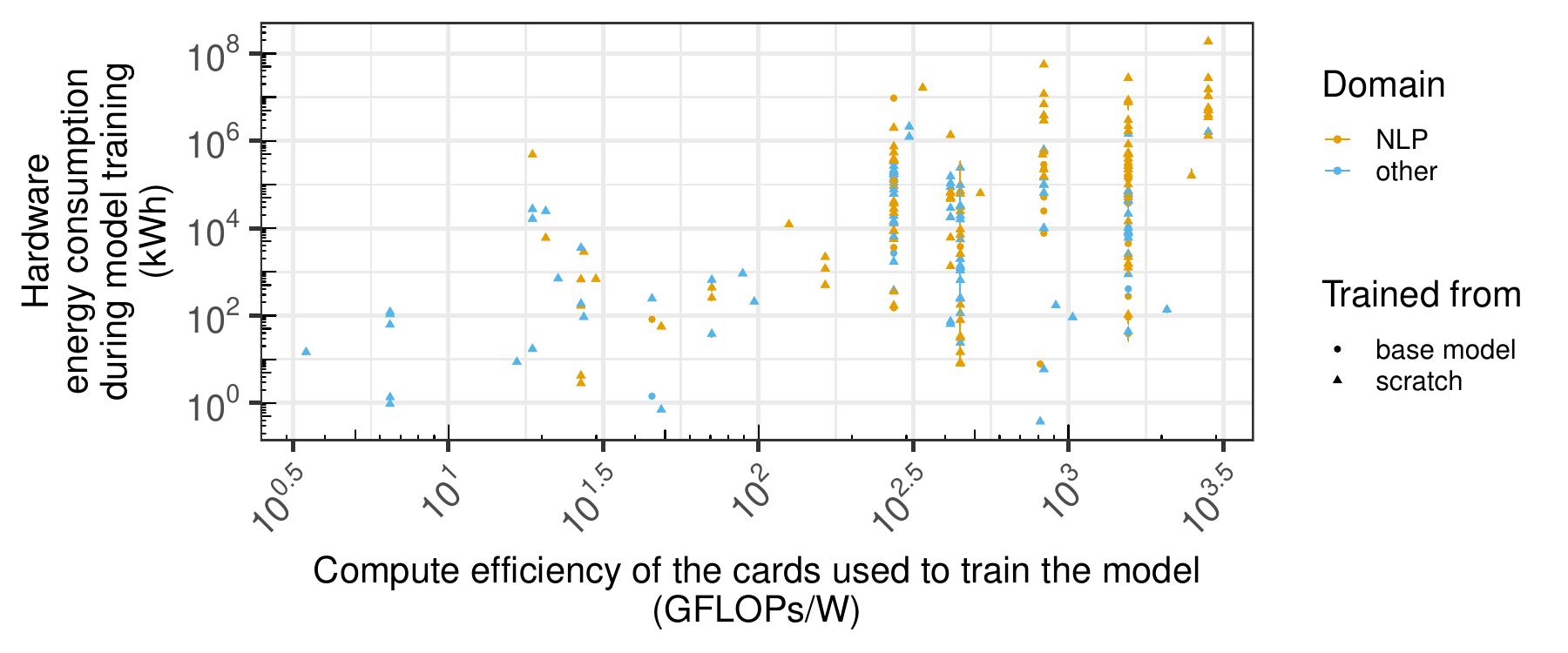}
    \caption{Hardware energy use during training over compute efficiency of the used card. Value intervals account for ambiguous card name}
    \label{fig:evol_consumption_over_efficiency}
\end{figure}

Figure~\ref{fig:evol_compute_efficiency} presents the evolution of the compute efficiency of the cards used to train \gls{ml} models over time. Compute efficiency has increased exponentially, meaning that hardware optimization has been leveraged extensively, and that jobs with a given compute demand require an exponentially decreasing energy use. However, these efficiency gains appear to have mostly been used to foster larger training jobs and not to mitigate or reduce energy use. Indeed, as shown in Figure~\ref{fig:evol_consumption_over_efficiency}, the use of more compute efficient hardware is correlated with an (exponential) increase in hardware energy use during training.

\subsection{Software Optimization}

More efficient model architectures are continuously developed and allow to attain a set model performance with less resources~\citep{Patterson2022plateau}. Be they sparser or more efficient in other manner, these new architectures and models should require less compute per parameter to attain a given performance. We approximate model \textit{software efficiency} as the average compute per parameter in the model training. Software efficiency is thus expressed in FLOP / parameter. 
We expect that more recent models have benefited from algorithmic optimization, and would therefore be more software efficient for a given performance target. It is however unclear how software optimisations fare in a context of increasing model performance.

\begin{figure}[ht]
    \centering
    \includegraphics[width=0.95\columnwidth]{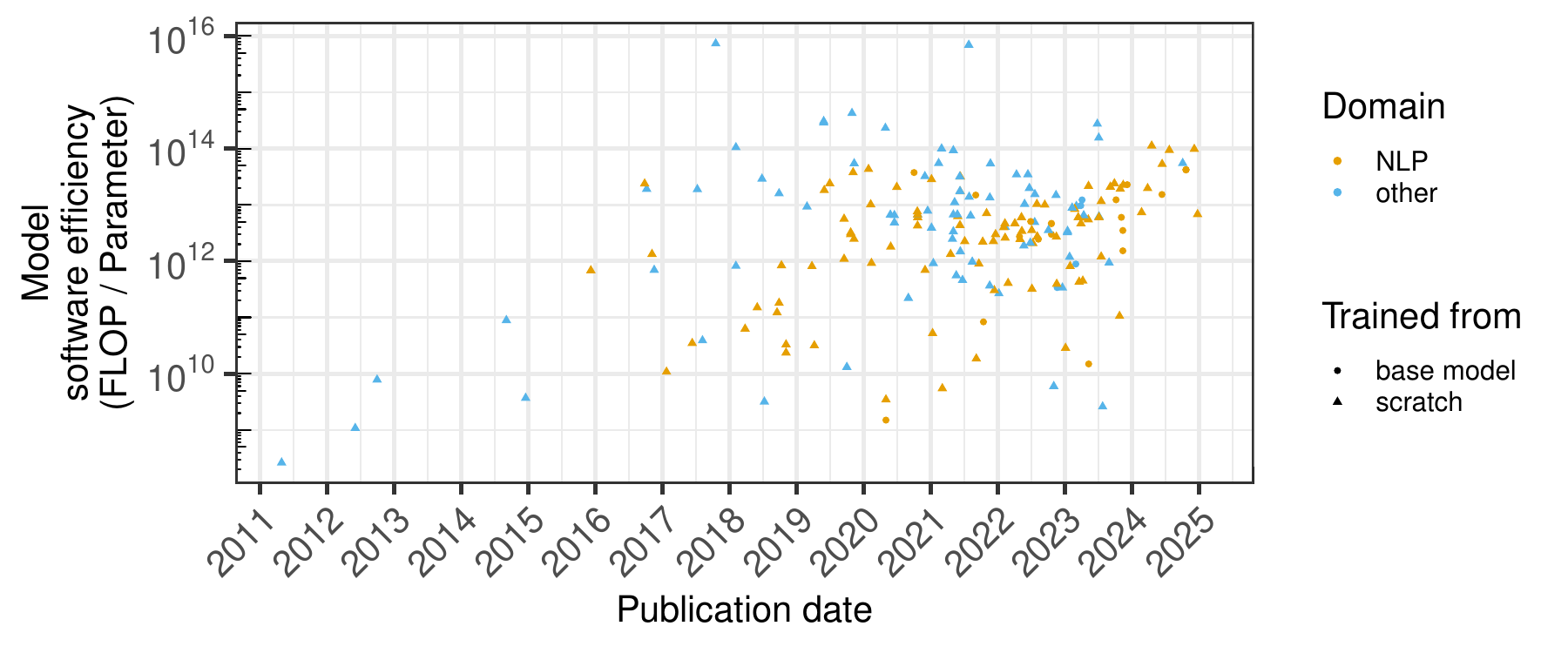}
    \caption{Software efficiency of models in Epoch AI over time. Lower is better. Y scale is logarithmic so a linear trend corresponds to an exponential growth.}
    \label{fig:software_efficiency}
\end{figure}

\begin{figure}[ht]
    \centering
    \includegraphics[width=0.95\columnwidth]{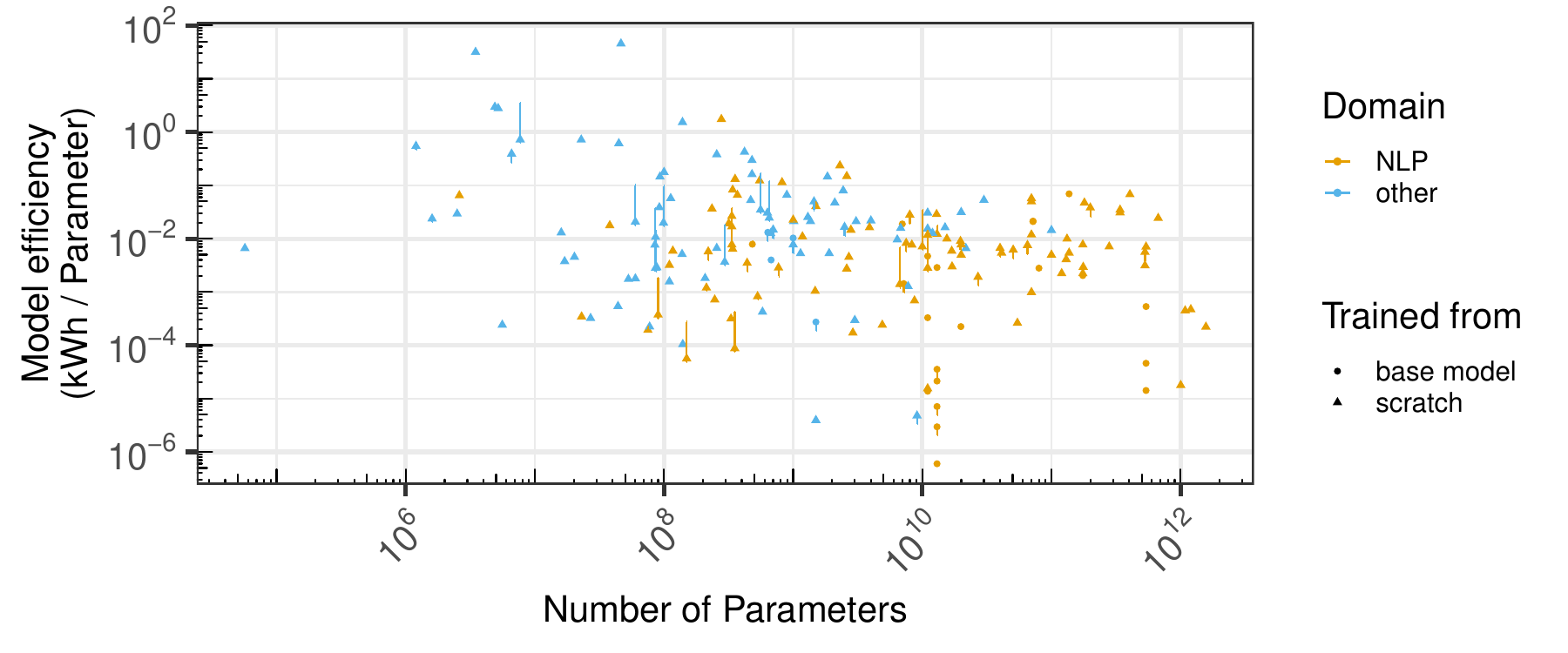}
    \caption{Combined efficiency of notable models in Epoch AI over model parameters. Lower is better. Value intervals account for ambiguous card name}
    \label{fig:combined_efficiency}
\end{figure}

Figure~\ref{fig:software_efficiency} details the evolution in software efficiency of notable models in the Epoch AI database over time. Software efficiency gains appear outweighed by increased compute demand to increase model performance, with a stable or even increasing software efficiency metric when we could have expected a decrease corresponding with the roll-out of the more efficient newer architectures. As for hardware optimization, our results suggest that software optimization have historically served model growth more than impact mitigation.

Hardware energy use during model training depends both on hardware and software characteristics. Figure~\ref{fig:combined_efficiency} details the combined efficiency of models training, expressed in kWh / parameter, over model size.
The combined efficiency appears to remain mostly constant with only a slight improvement with model size.

Hardware and software optimizations yield economies of scale that are outpaced by the increase in compute demand, resulting in a net increase in energy use for \gls{ml} training.

\subsection{Carbon optimization can only yield short-lived benefits}

\begin{figure}[ht]
    \centering
    \includegraphics[width=.95\columnwidth]{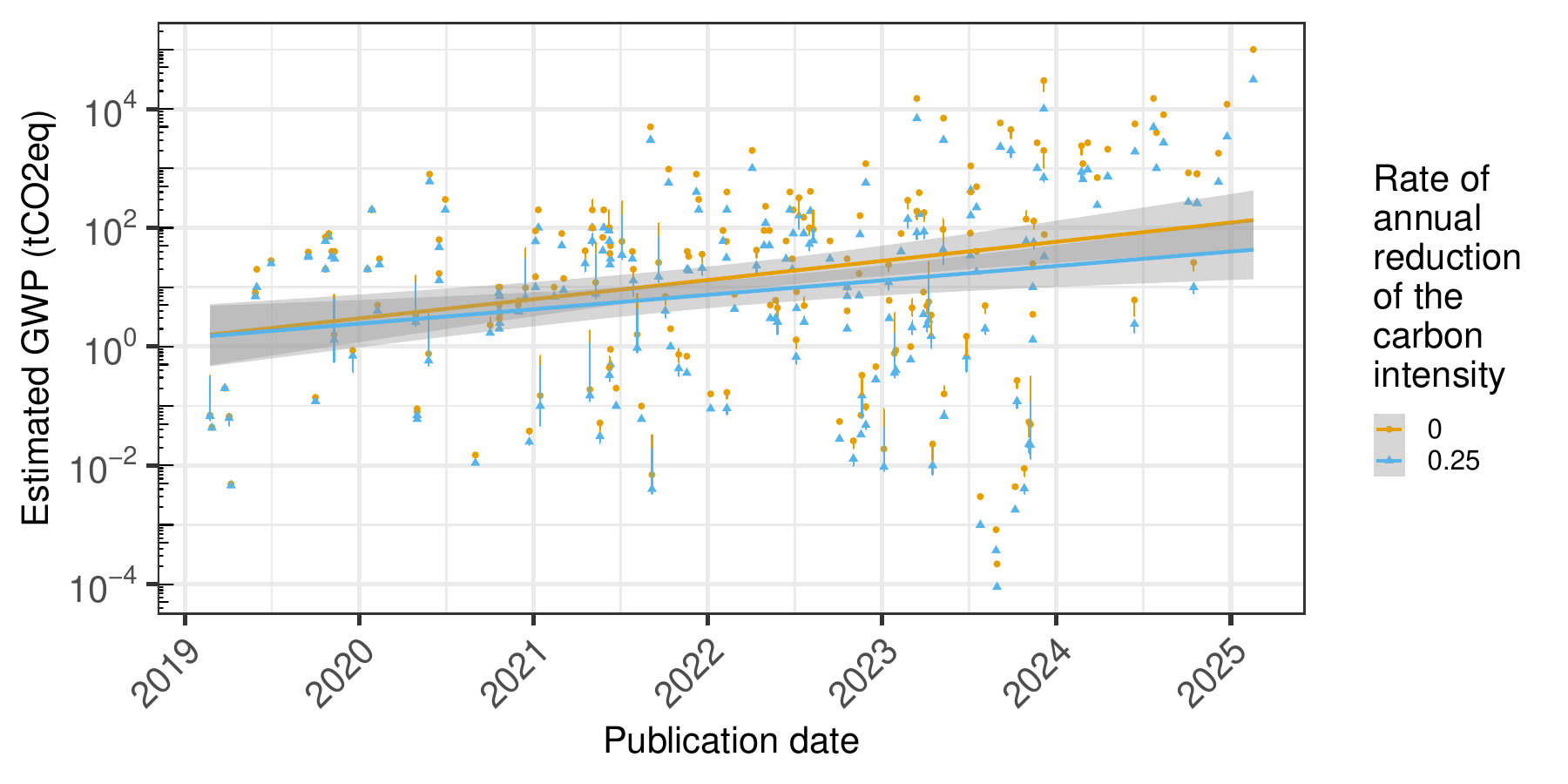}
    \caption{Estimated carbon footprint of training models released after 2019, with or without reduction of the \gls{CI} of the used electricity.}
    \label{fig:reduction_CI}
\end{figure}

We compare the impact of models trained with actual electricity mixes with a simulated annual reduction of 25\% of the \gls{CI} of electricity mixes (see Section~\ref{subsec:carbon_reduction}).
Figure~\ref{fig:reduction_CI} shows that the carbon footprint of models released from 2019 is increasing regardless of carbon optimizations.
Trends are very similar with or without \gls{CI} reduction and regression coefficients are significantly positive. 
This suggests that carbon optimization is not a sufficient reduction strategy.

\gls{CI} for electricity is limited by the lowest global mixes (15-20g\COtwo{}/kWh), meaning that even fast \gls{CI} reductions cannot keep pace with rising energy use. 
While using low-impact electricity is a key Green AI practice, it cannot alone curb the growth of training impacts. Furthermore, this strategy fails to address the increasing environmental costs of hardware production. 
Worse, shifting data centers geographically to reduce \gls{CI} may incentivize shorter lifespan for these facilities, further increasing their total environmental impact.

\section{Discussion}
\label{sec:Limitations-Discussion}

To our knowledge, this is the first study proving the growth in \gls{ml} training impacts and explaining drivers of this growth. We show that carbon optimization strategies fail 
to curb training impacts.

\subsection{Impact Shifting}

Computing facilities are increasingly energy efficient and can perform much more computation in a given duration using less energy. This efficiency is partly obtained by regular hardware updates in data centers. Replacement can occur at the end-of-life of hardware or be more premature, shortening the lifespan of the devices. Frequently changing hardware could be understood as a form of \textit{impact shifting}. Energy use can be reduced (at constant use) during the potentially shorter equipment lifespan, so usage impact is reduced. However, production impact in terms of carbon footprint and metallic resource depletion is increased by frequent hardware renewal. In addition, new hardware also has higher production impact due to technological advances required to reduce energy use in the usage phase.

\subsection{Rebound Effect is Prevalent in AI} Our results suggest rebound effect, consistently with previous observations: 
\citet{Wu2022sustainable} identify a rebound effect at Facebook. \citet{Patterson2022plateau} show that a constant share of Google's energy use is attributable to AI when the total use increased. 
Efforts from \citet{devries2023growing} and \citet{Desislavov2023trends} suggest that our findings should also apply to inferences. 

Hardware and algorithmic optimization appear to have fueled increases in compute demand to chase better model performances. Their role thus appears not so much as a mitigation but as an enabler of growing impacts.

\subsection{Greener Energy Cannot Void Carbon Impact}
Figure~\ref{fig:reduction_CI} shows that reducing \gls{CI} of electricity used seems insufficient to curb the exponential growth of the carbon footprint of training \gls{ml} models. Furthermore, the electricity use of data centers destabilizes local electricity grids \citep{Ortar2023powering}, potentially causing the prolongation of fossil fuel power plants \citep{FT2024coal, Guardian2024Ireland}. On-site renewable energy production to match data centers use can also be a source of electricity grid instability \citep{Gnibga2024renewable}. Matching the carbon footprint 
through carbon offsetting also has limited potential~\citep{Guardian2023carbon,Lohmann2009toward}.

\subsection{Community effort to document environmental impact is required}

\begin{table*}[t]
    \centering
    \begin{tabular}{cc}\toprule
        Information & Details \\\midrule
        Training duration & time\\
        Card model & name \\
        Number of cards & number \\
        Server model and configuration & name and (number CPU, memory volume\dots)\\
        Type of datacenter & e.g., laboratory facilities, hyperscaler,... \\
        Hardware usage during training & ratio \\
        Datacenter efficiency & typically \gls{PUE}\\
        Datacenter location & country or state\\\bottomrule
    \end{tabular}
    \caption{Proposed standard reporting format to ensure sufficient information for environmental assessment of final model training}
    \label{tab:proposed_standard}
\end{table*}

The \gls{nlp} community has identified the need to rethink progress and evaluation practices. Environmental issues surrounding \gls{llm} (at all the stages of their life cycles) are significant, making it pressing to include environmental impacts in \gls{nlp} evaluation.
Sole reporting of carbon footprint figures are insufficient as environmental assessments are complex and currently often difficult to reproduce \citep{Morand2024empreinte}.
More transparency is needed, ensuring that papers report all the necessary information to produce environmental assessments, notably including hardware used, training duration, compute localisation.
A standardised reporting format could be proposed for instance in the form of a Table, or cards such as proposed by \citet{hershcovich-etal-2022-towards}, \citet{chiu2026position} or \citet{Henderson2020towards}.
Table~\ref{tab:proposed_standard} details the necessary information for model training in the form of a proposed standard table.

\section{Conclusion} 
\label{sec:conclusion}

We show that the environmental impacts of training language models, and \gls{ml} models in general, grew exponentially from 2013 to 2025. Despite hardware, algorithmic and carbon optimizations, impacts keep growing, consistent with rebound effect. This growth of impacts is only part of the environmental dynamics of AI. It does not account for impact sources other than individual model training, such as inference.
Obtaining information to plot the trends of training impact is a grinding task that could be alleviated by a community wide effort to document environmental impacts as part of NLP evaluation. Integrating environmental impacts in NLP evaluation would give better visibility of impacts to the community and support the use of impact as a feature in research planning and decision making. 
Impact reduction must be combined with a broader consideration of the role of AI in a sustainable society.

\section{Limitations}

\subsection{Study Scope Supports Reliable Trends} 

Production impact assessment in MLCA has limitations, including not accounting for technological node and assuming constant memory density, but estimates are consistent with recent research~\citep{schneider2025lifecycleemissionsaihardware}. The use of the PUE may lead to underestimation of infrastructure use.
Adjusting for dynamic ratio as proposed by~\citet{Morand2024MLCA}
would not change the overall findings since it would only mean using a higher multiplicative factor for all models. Epoch AI, although extensive, does not include all models and variants, but omissions are unlikely to alter the observed exponential increase in training impacts.

\subsection{Sparse Information Still Leads to Consistent Estimations}

This study estimates missing parameters for models using inferences and hypotheses, which may lead to under or over-estimation of certain parameters (e.g., training time based on number of FLOP). The validity of these estimates is checked by comparing our estimations to published information. These check yield to some significant differences but general consistency. For example, the carbon footprint for the BLOOM-176B model presents one of the greatest difference: our model estimates a carbon footprint of 220t\COtwo{} instead of 50-80t\COtwo{}, due to assumptions about its training location and electricity mix. Yet, most other models have close or very close estimations, as for the case of the GPT-4 model used as an example for our methodology where our model estimation of 15kt\COtwo{} is very close to the reference estimation of 12 to 15 15kt\COtwo{}\footnote{The reference estimation only accounts for energy use and not hardware production.}.

\subsection{This Study Addresses Model Training} 

\citet{morrison2025holistically} and \citet{Luccioni2022estimating} have 
shown that model development includes training multiple smaller models to converge on final architecture. These multiple training runs, that we do not account for, can account for at least half of the total footprint~\citep{morrison2025holistically}.
Future work can also address the impact of inferences in the sector, as they weight more heavily over the life cycle of models. It will however require access to user information.

\subsection{Impacts Go Beyond Carbon Footprint and Resource Depletion}
\label{subsec:discussion-metrics}

This study examined the carbon footprint and metallic resource depletion of training AI models, finding that both metrics have increased over time due to impacts in different phases in the life cycle of hardware. Importantly, AI environmental impacts extend beyond these metrics, including water usage~\citep{Mytton2021datacenter}, ecosystem destruction~\citep{Comber2023Computing}, and pollution from hardware mining and disposal. Additionally, AI poses significant social consequences and ethical challenges~\citep{Bender2021parrots,Jiang2023Art}, highlighting the need for a more comprehensive assessment that incorporates qualitative analyses and a broader range of impacts.

\section*{Ethical Considerations}

This study provides data for the footprint of training \gls{ml} models, including \gls{llm}, over a decade using estimations and information from the EpochAI database and GPU manufacturers. No re-training of the models was needed to compute impact metrics. 

\section*{Acknowledgments}

This work has received funding from the French "Agence Nationale pour la Recherche" under grant agreement InExtenso - ANR-23-IAS1-0004. 
Clément Morand was supported by a doctoral grant from ENS Rennes.  
The authors thank TechPowerUp for allowing us to use their data in this study. 
Clément Morand would like to thank Aina Rasoldier for his help in
gathering the data from TechPowerUp, Loïc Lannelongue for discussions
during the experiments framing and Adrien Berthelot for insightful discussion and bibliographical advice. 
The authors are grateful to Aurélie Bugeau, Karën Fort, Caio Corro and Pierre Zweigenbaum, as well as the anonymous reviewers, for their comments on earlier versions of this manuscript.

\bibliography{custom}

\appendix

\section{Details for Exact Replication of Environmental Assessments}
\label{app:replication}

\subsection{Processing Ambiguous Card Names}

8 \gls{ml} models are excluded from analysis because various hardware is used throughout the training process and adequate attribution is not possible. 
Other models document hardware ambiguously (e.g., "A100" could refer to several cards with different features).  
Then, the one we expect to be most frequently used, is chosen as the reference value, and other options are used to compute a value interval.

\paragraph{Server Characteristics}

We made hypotheses on memory provisioning and number of cards per server, and hardware lifespan and utilization using values consistent with computing facilities at our university, documentation from graphic card and server manufacturers and the literature~\cite{Ostrouchov2020gpu,Wu2022sustainable}. Exact details for replication on server characteristics is available in \ref{sub-app:servers}. This information serves to allocate server production impact to a specific model training. Allocated production impacts are called \emph{embodied impacts}.

We also modeled the increase in data center efficiency estimating average PUE from 2010 to 2018  using linear interpolation and a PUE of 1.2 from 2018 onward based on \citet{Masanet2020recalibrating}. 

\paragraph{Hardware Energy use.}

We use a value of 100\% usage both for GPUs and CPUs during training as it was shown to yield accurate estimations~\cite{Jay2023experimental}.

\paragraph{Environmental Impact of Energy Usage}

To evaluate the impacts of energy use associated with training each model, we consider that models have been trained using the energy mix of the countries
of the \gls{ml} model producers 
documented by Boavizta.\footnote{
Details on the sources used for each countries can be found at: \url{https://github.com/blubrom/MLCA/blob/main/boaviztapi/data/electricity/electricity_impact_factors.csv}.} If multiple countries participate in model creation, the energy mixes of each implicated country are used to create a value interval, and the mix of the country indicated in first position is used as the reference value.

\subsection{Training Duration Estimates}
\label{sub-app:duration}

\paragraph{Ensuring Consistency and Excluding Anomalies:} In order to validate estimates from \secondestimate{} and correct the underestimation, we compare \firstestimate{} and \secondestimate{} for each of the  119 models where both values are available, taking \firstestimate{} as a ground truth.
This allows us to obtain an estimate for the ratio of hardware performance: $\text{ratio} = \frac{\secondestimate{}}{\firstestimate{}}$. This ratio should in theory be lower than one as a ratio greater than one would in theory mean that the model has used the GPUs at more than their peak performance.
We define \textit{anomalies} as models where the obtained ratio is greater than one or smaller than 10\%. Analyzing the obtained ratios revealed anomalies for 19 models (in 16\% of cases).
Anomalies especially occur with cases of fine-tuned models where \secondestimate{} includes training the base models while \firstestimate{} only accounts for the fine-tuning process.
Anomalies also occur when the number of FLOP in the EpochAI database was estimated based on compute power values different from the ones in our graphics cards database due to inconsistencies between the Epoch AI database and manufacturer data. Both types of anomalies should not pose problems as we use \firstestimate{} as our final estimation for these models.

\paragraph{Linear Model for Correcting Underestimations:} After checking for consistency between the two estimation methods and excluding benign anomalies, we develop a linear model to correct the underestimation from \secondestimate{} compared to \firstestimate{}.
We build a linear model to predict \firstestimate{} using \secondestimate{}. This model is computed on 100 observations excluding the anomalies. 
To ensure a linear relation between both variables, we estimate $\log(\firstestimate{}) \sim \log(\secondestimate{})$  as \firstestimate{} and \secondestimate{} both are exponentially distributed.
The linear regression analysis revealed a statistically significant model (F(1,98) = 6525, p $< 2 \times 10^{-16}$), with an adjusted R² of 0.98, meaning that 98 percent of the variance in the observations is explained by our model. 
The model equation is $\log(\firstestimate{}) = 1.3 + 1.0 \log(\secondestimate{})$ with a standard error of $0.12$ for the intercept and $0.01$ for the regression coefficient. This indicates that an increase of 1 for the $\log(\secondestimate{})$ value leads to an average increase of 1.00 units in $\log(\firstestimate{})$. This positive relationship between $\log(\secondestimate{})$ and $\log(\firstestimate{})$ was found to be statistically significant (t(98) = 80.78, p $< 2 \times 10^{-16}$), affirming the predictive power of $\log(\secondestimate{})$ on $\log(\firstestimate{})$.
In addition to the regression analysis, a scatterplot with the fitted regression line was examined to ensure model assumptions were met. Homoscedasticity was confirmed (studentized Breusch-Pagan test = .08, p = 0.77) and the residuals appeared to be independent (Durbin-Watson D = 1.81, p = .17).

\subsection{Details on Modeled Server Characteristics}
\label{sub-app:servers}

We made the hypothesis that servers equipped with NVIDIA workstations cards contain 4 graphics cards, 2 CPUs and 512 GB memory. We also made the hypothesis that servers equipped with NVIDIA non-workstations cards contain 2 graphics cards, 2 CPUs and 192 GB memory. 
These configurations were chosen based on a review of configurations in the different generations of NVIDIA DGX servers as well as on configurations in different computing facilities, including computing facilities available in the Université Paris-Saclay.

For non-NVIDIA hardware, we searched documentation (e.g., Google Cloud Platform documentation, publications by Google) to obtain information on the number of chips and processors per server. For instance, for TPUv3, a server with two CPUs manages every four TPU chips. Missing information on the memory quantity in each server, the value of 448 GB memory is chosen based on the value for the TPU v5p for the TPU v2, v3 and v4. This value seems consistent with the value chosen for the NVIDIA servers.

We use values consistent with hyper-scale data centers for the hardware lifespan, average utilization over its life cycle and infrastructure energy use. We use a lifespan of 3 years for the hardware \citep{Ostrouchov2020gpu}, and, using information from Meta, a hardware utilization of 50\%~\citep{Wu2022sustainable}.

Based on the data from \citet{Masanet2020recalibrating}, we model an increase in the efficiency of data centers between 2010 and 2018, accompanied with the shift from traditional data centers to hyperscale data centers.
Assuming a uniform distribution between Traditional, Cloud and Hyperscale data centers, we use an average PUE of 1.75 for data centers in 2010. (This average value of 1.75 is close to the average PUE value for Cloud (non-hyperscale) data centers in 2010)
Assuming that all AI workloads take place in hyperscale data centers starting in 2018, we use an average PUE of 1.2 from 2018 onwards.
For dates between 2010 and 2018, we use a linear interpolation to estimate the used average PUE.

\subsection{Software Used in the Analysis}

Data manipulation and statistical analyses have been performed using emacs Org mode 9.1.9, Python 3.8.10 using the pandas version 2.0.3 library and R version 3.6.3 (2020-02-29) with the ggplot2\_3.4.3,  dplyr\_1.1.3, lmtest\_0.9-40, stringr\_1.5.0 and zoo\_1.8-12  libraries.  Platform: x86\_64-pc-linux-gnu (64-bit) Running under: Ubuntu 20.04.6 LTS

\section{Regression analyses of the observed trends}
\label{app:regression}

\subsection{Training Energy use over time}

We build a linear model to predict model training energy use (in Wh) using the model release date. To ensure a linear relation between both variables, we estimate $\log(\text{model training energy use}) \sim \text{model release date})$  as model training energy use is exponentially distributed. We use weighted least square regression to account for heteroscedasticity due to increasing variance in model training energy use in more recent years. 
The linear regression analysis revealed a statistically significant model (F(1,221) = 105, p $< 2.2 \times 10^{-16}$), with an adjusted R² of 0.32, meaning that 32 percent of the variance in the observations is explained by our model. 
The model equation is $\log(\text{model training energy use}) =  -1.49e^{+01} + 8.58e^{-04} \times \text{release date}$ with a standard error of $ 1.54$ for the intercept and $8.38e-05$ for the regression coefficient. This indicates that an increase of 1 for the release date value leads to an average increase of 8.582e-04 units in $\log$(model training energy use). This positive relationship between release date and model training energy use was found to be statistically significant (t(221) =  10.25, p $< 2 \times 10^{-16}$), affirming the predictive power of release date on model training energy use.

\subsection{Training carbon footprint}

We build a linear model to predict model training carbon footprint (in kg\COtwo{}) using the model release date. To ensure a linear relation between both variables, we estimate $\log(\text{model training carbon footprint}) \sim \text{model release date})$  as model training carbon footprint is exponentially distributed. We use weighted least square regression to account for heteroscedasticity due to increasing variance in model training carbon footprint in more recent years. 
The linear regression analysis revealed a statistically significant model (F(1,221) = 117.7, p $< 2.2 \times 10^{-16}$), with an adjusted R² of 0.35, meaning that 35 percent of the variance in the observations is explained by our model. 
The model equation is $\log(\text{model training carbon footprint}) =  -1.62e^{+01} + 9.11e^{-04} \times \text{release date}$ with a standard error of $1.55$ for the intercept and $8.39e-05$ for the regression coefficient. This indicates that an increase of 1 for the release date value leads to an average increase of $89.11e^{-04}$ units in $\log$(model training carbon footprint). This positive relationship between release date and model training carbon footprint was found to be statistically significant (t(221) =  10.85, p $< 2 \times 10^{-16}$), affirming the predictive power of release date on model training carbon footprint.

\subsection{Training metallic resource depletion}

We build a linear model to predict model training metallic resource depletion (in kg\Sbe{}) using the model release date. To ensure a linear relation between both variables, we estimate $\log(\text{model training metallic resource depletion}) \sim \text{model release date})$  as model training metallic resource depletion is exponentially distributed. We use weighted least square regression to account for heteroscedasticity due to increasing variance in model training metallic resource depletion in more recent years. 
The linear regression analysis revealed a statistically significant model (F(1,221) =  94.31, p $< 2.2 \times 10^{-16}$), with an adjusted R² of 0.30, meaning that 30 percent of the variance in the observations is explained by our model. 
The model equation is $\log(\text{model training metallic resource depletion}) =  --1.59e^{+01} + 8.10e^{-04} \times \text{release date}$ with a standard error of $1.54$ for the intercept and $8.34e^{-05}$ for the regression coefficient. This indicates that an increase of 1 for the release date value leads to an average increase of $8.10e^{-04}$ units in $\log$(model training metallic resource depletion). This positive relationship between release date and model training metallic resource depletion was found to be statistically significant (t(221) =  9.71, p $< 2 \times 10^{-16}$), affirming the predictive power of release date on model training metallic resource depletion.

\section{Sensitivity analysis on the estimated GPU average performance during training}
\label{app:sensitivity}

As detailed in Section \ref{subsec:Machine-learning} and Appendix \ref{sub-app:duration}, training duration estimates using \secondestimateclean{} depend on an empirically derived assumption about average GPU performance during the training process. Our model corresponds to a quasi-constant performance ratio of around $27~\%$. This section details a sensitivity analysis examining the impact of variations in the used performance ration on the obtained results and trends.

\begin{figure}[ht]
    \centering
    \includegraphics[width = \linewidth]{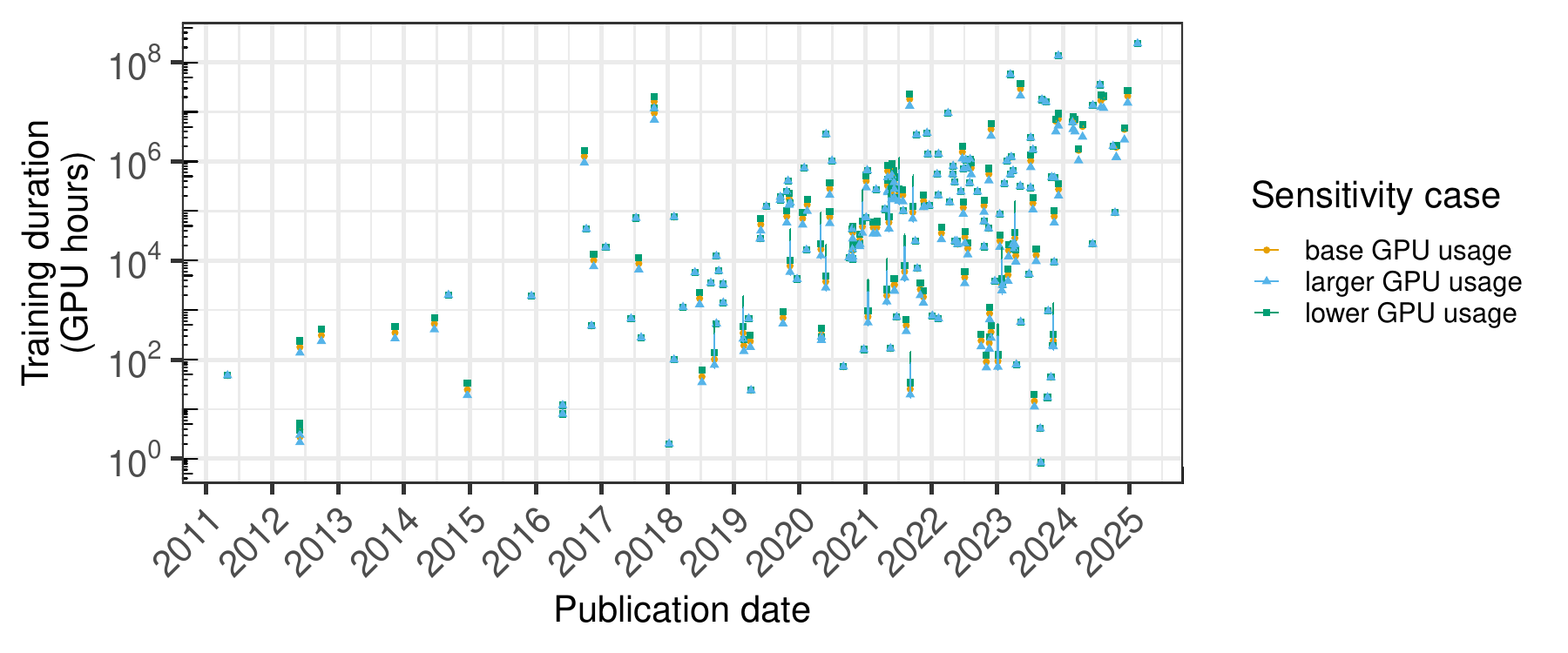}
    \caption{Estimated training duration for models in Epoch AI over time in the three considered graphics card performance scenarios.}
    \label{fig:sensitivity_GPU-hours}
\end{figure}

We tested two variations corresponding to a lower usage of the cards of $20~\%$ and a larger usage of the cards of $35~\%$. Larger usage means that the cards are more efficient at using their compute power, corresponding to shorter training duration. Figure~\ref{fig:sensitivity_GPU-hours} details the estimated training duration in the different performance scenarios. As expected, slightly shorter training duration are  estimated when considering a larger usage of the cards ($-16.5~\%$ in average, compared to the base performance ratio), and slightly longer training duration are estimated when considering a lower usage of the cards ($+10.2~\%$ in average, compared to the base performance ratio). The change in hypothesis does not affect the ($56~\%$ of) models where \firstestimate{} is used. This variation in estimated training duration leads to slight variations in the final estimated training footprints (respectively $-16.7~\%$ and $+10.1~\%$ in average \gls{GWP}, compared to the base estimated \gls{GWP} and espectively $-18.1~\%$ and $+9.5~\%$ in average \gls{ADPe}, compared to the base estimated \gls{ADPe}). These variations do not change conclusions on the observed trends.

\printnoidxglossaries

\end{document}